\documentclass[letterpaper, 10 pt, conference]{ieeeconf}  

\IEEEoverridecommandlockouts                              

\usepackage{amsmath} 
\usepackage{amssymb}  
\usepackage{graphicx}
\usepackage[adjust]{cite}

\usepackage{pgfplots}
\pgfplotsset{compat=1.8}
\usepackage{tikzscale}

\usepackage{xcolor}
\usepackage{xspace}
\usepackage{csquotes}

\usepackage{caption}
\usepackage{subcaption}

\usepackage{algorithm}
\usepackage[noend]{algpseudocode}
\algrenewcommand\algorithmicrequire{\textbf{Given:}}
\algrenewcommand\algorithmicensure{\textbf{Wanted:}}

\usepackage{mathabx}

\usepackage{siunitx}

\usepackage{paralist}

\usepackage[hidelinks]{hyperref}

\title{\LARGE \bf
Show me what you want: Inverse reinforcement learning to automatically design robot swarms by demonstration
}

\author{Ilyes Gharbi$^{1}$, Jonas Kuckling$^{1}$, David Garzón Ramos$^{1}$, and Mauro Birattari$^{1}$
\thanks{$^{1}$IG, JK, DGR, and MB are with IRIDIA, Université libre de Bruxelles, Brussels, Belgium. {\tt\small \{ilyes.gharbi, jonas.kuckling, david.garzon.ramos, mauro.birattari\}@ulb.be}}%
\thanks{The project has received funding from the European Research Council (ERC) under the European Union’s Horizon 2020 research and innovation programme (Demiurge: grant agreement No 681872) and from Belgium’s Wallonia-Brussels Federation through a ARC Advanced Project 2020 (Guaranteed by Optimization). JK and MB acknowledge support from the Belgian Fonds de la Recherche Scientifique-FNRS. DGR acknowledges support from the Colombian Ministry of Science, Technology and Innovation – Minciencias.}
\thanks{The experiments were designed by IG, JK, and DGR and performed by IG and JK. The paper was drafted by IG and JK and edited by IG, JK, DGR, and MB; all
authors read and commented the final version. The research was directed by MB.}%
}

\newcommand{\cho}{\texttt{Choc\-o\-late}\xspace}

\newcommand{\stick}{\texttt{EvoStick}\xspace}
\newcommand{\epuck}{e-puck\xspace}

\newcommand{\argos}{ARGoS3\xspace}

\newcommand{\DemoCho}{Demo-\texttt{Cho}\xspace}
\newcommand{\OEvo}{O-\texttt{Evo}}
\newcommand{\OCho}{O-\texttt{Cho}}
\newcommand{\DCho}{D-\texttt{Cho}}

\newcommand{\width}{0.23\textwidth}
\newcommand{\heigth}{0.33\textwidth}

\begin{document}

\maketitle
\thispagestyle{empty}
\pagestyle{empty}

\begin{abstract}
Automatic design is a promising approach to generating control software for robot swarms.
So far, automatic design has relied on mission-specific objective functions to specify the desired collective behavior.
In this paper, we explore the possibility to specify the desired collective behavior via demonstrations.
We develop \DemoCho, an automatic design method that combines inverse reinforcement learning with automatic modular design of control software for robot swarms.
We show that, only on the basis of demonstrations and without the need to be provided with an explicit objective function, \DemoCho successfully generated control software to perform four missions.
We present results obtained in simulation and with physical robots.
\end{abstract}

\section{INTRODUCTION}

\noindent Swarm robotics is an approach to control large groups of autonomous robots \cite{Ben2005sab,Sah2005sab,DorBirBra2014SCHOLAR}.
It is considered a prominent research direction \cite{YanBelDup-etal2018SCIROB} and has attained a notable position in the literature \cite{RubCorNag2014SCI,WerPetNag2014SCI,GarBir2018SCIROB,SlaCarCar-etal2018SCIROB,YuWanDu-etal2018NATUCOM,LiBatBro-etal2019NATU,XieSunFan-etal2019SCIROB,HasLigRudBir2021NATUCOM}.
A robot swarm is a decentralized system and consists of relatively simple robots that can perceive and interact with the environment only in their local neighborhood.
A swarm is a self-organizing system, that is, its collective behavior emerges from the interactions of its individual robots.
The design challenge in swarm robotics is to program the individual robots so that a desired collective behavior emerges.
Several methods have been proposed for specific classes of missions \cite{HamWor2008SI,Kaz2009IJICC,BerKumNag2011icra,BeaDulUsb-etal2012fpadl,BraBruDorBir2014ACMTAAS,ReiValFer-etal2015PLOSONE,LopTreLea-etal2016SI,PinBel2016IEEESW,Ham2018book}. Yet, due to the many unpredictable interactions within the swarm, no generally-applicable and principled method exists to design a desired collective behavior \cite{BraFerBirDor2013SI,DorTheTri2020SCIROB,DorTheTri2021PIEEE}.


Automatic off-line design has proven to be a viable approach for the design of control software for robot swarms \cite{BalTriBon-etal2007IEEETSMCB,GauCheLi-etal2014IJRR,FraBraBru-etal2014SI,DuaCosGom-etal2016PLOSONE,JonStuHauWin2018dars}---other related approaches exist \cite{FraBir2016FRAI,BreHaaPri2018FRAI,BirLigHas2020NATUMINT}.
In automatic off-line design, an optimization algorithm searches the space of possible instances of control software to find one that maximizes a given mission-specific objective function, which measures the performance of the swarm.
The objective function is typically assessed through simulations.
The selected instance of control software is then uploaded to real robots, which are then deployed in the target environment to perform the mission.
Notably, no human intervention beyond the specification of the mission takes place \cite{BirLigHas2020NATUMINT}.
%
%
The objective function is part of the formal specification of the mission at hand.
Defining an objective function is challenging, and requires to be familiar with mathematical modeling.
This is a task that requires the attention of a skilled professional and could not be performed by an end user.
%

The problem of defining an appropriate objective function is similar to the problem that in the reinforcement learning literature goes by the name of \emph{reward shaping}: the definition of a reward function that facilitates learning a desired policy~\cite{NgHarRus1999icml}.
Inverse reinforcement learning is an approach to address this problem: instead of learning a policy that maximizes a given reward function, inverse reinforcement learning algorithms learn a reward function from demonstrations of an optimal behavior.
The learned reward function can then be used to generate a policy that reproduces the demonstrated behavior.
Inverse reinforcement learning is motivated by the fact that, for some classes of problems, demonstrating an optimal behavior is easier than defining a properly shaped reward function~\cite{Rus1998colt,AbbNg2004icml}.
One of the earliest proposed approaches to inverse reinforcement learning is \emph{apprenticeship learning}~\cite{AbbNg2004icml}.
Given demonstrations of the desired behavior, the apprenticeship learning algorithm iterates between \begin{inparaenum}[i)]\item learning a policy based on an intermediate reward function and \item learning a new intermediate reward function based on the behavior of the previously generated policies\end{inparaenum}.
The algorithm stops if the behavior of the current policy is sufficiently close to the provided demonstrations.

We contend that inverse reinforcement learning can be adopted in the framework of the automatic design of control software for robot swarms: instead of defining a mission-specific objective function, we can provide demonstrations of the desired swarm behavior and let an inverse reinforcement learning algorithm infer an objective function to automatically generate the control software that produces the desired behavior itself.
In this work, we focus on desired behaviors that can be described through the final position of the robots.

\section{RELATED WORK}
\label{sec:related-work}

\noindent Inverse reinforcement learning has already found application in robotics: Krishnan et al. proposed SWIRL, an inverse reinforcement learning algorithm to learn various robot tasks, including parallel parking and surgical cutting along a line \cite{KriGarLia-etal2019IJRR}.
The robot successfully learned the tasks from demonstrations and the learned policies were robust to perturbations, such as different initial positions.

Inverse reinforcement learning was also studied in the scope of multi-agent systems. 
Natarajan et al. used inverse reinforcement learning to develop a centralized controller that coordinates multiple traffic lights \cite{NatKunJud-etal2010icmla}.
Song et al. used inverse reinforcement learning to design policies in general Markov games \cite{SonRenSad-etal2018ANIPS}.

In swarm robotics, Šošic et al. used inverse reinforcement learning to learn swarm policies from trajectories obtained from simulations of two particle models \cite{SosKhuZou-etal2016arxiv}.
The results show that the swarm was able to replicate the behavior of both particle models.
However, the design process required the complete behavior to be already pre-implemented so as to serve as a demonstration.

Besides inverse reinforcement learning, other approaches have been adopted in swarm robotics to learn collective behaviors from demonstrations.
Li et al. proposed Turing learning, a method that enables robots to imitate the behavior of other pre-programmed robots, without the need to manually specify the set of features that describe the desired behavior \cite{LiGauGro2016SI}.
However, the approach assumes that an implementation of the desired behavior exists and can be used to generate demonstrations.
Alharthi et al. extracted swarm behaviors from video demonstrations and used evolutionary algorithms to synthesize control software in the form of behavior trees \cite{AlhAbdHau2022arxiv}.
Also in this case, the approach requires that an implementation of the desired behavior exists.

\section{APPRENTICESHIP LEARNING}
\label{sec:apprenticeship-learning}

\noindent Reinforcement learning problems are commonly modelled as a Markov decision process $M=(S,A,T,\gamma,R)$ \cite{KaeLitMoo1996JAIR}.
A reinforcement learning algorithm learns a policy $\pi:~S~\rightarrow~A$ that maximizes the expected sum of discounted rewards: $E_{s_0}[V_M^\pi(s_0)]=E_{s_0}[\sum_{t}{\gamma^t R(s_t)|\pi]}$, with $s_0, \dots, s_t \in S$.

In inverse reinforcement learning, the reward function $R$ is not provided.
Instead, demonstrations of the desired behavior are given in the form of sequences of states.
It is assumed that a \enquote{true} reward function $R^*$ exists and it is such that the policy $\pi^*$ that maximizes the value function based on $R^*$ would generate the given demonstrations.

In apprenticeship learning \cite{AbbNg2004icml}, it is furthermore assumed that there exists some mapping $\phi: S \rightarrow [0,1]^k$ that maps the states of the system to a $k$-dimensional vector of features.
The \enquote{true} reward function $R^*$ is assumed to be a linear combination of the features: $R^*(s)=w^* \cdot \phi(s)$, where $w^*~\in~\mathbb{R}^k$ and $s \in S$.
For every policy $\pi$, a feature expectation can be defined as $\mu(\pi)=E_{s_0}[\sum_{t}{\gamma^t \phi(s_t)}|\pi]~\in~\mathbb{R}^k$.
It follows that, for $R^*$, $E_{s_0}[V_M^\pi(s_0)]=w^* \cdot \mu(\pi)$.
When the expectation cannot be computed formally, it can be replaced by an empirical estimate $\hat{\mu}(\pi)$ computed on the basis of sampled trajectories.
With $\mu_E$, we indicate the feature expectation of the provided demonstrations.

Algorithm \ref{alg:apprenticeship-learning} shows the pseudo-code of the apprenticeship learning algorithm.
Given the mapping $\phi$ and the feature expectation $\mu_E$ of the demonstrations, the algorithm iteratively refines the vector of weights $w$, until the observed feature expectation $\mu_i$ approximates $\mu_E$.
At every iteration, a support vector machine \cite{HeaDumOsu-etal1998IEEEISA} is fitted on $\mu_E$ and all encountered $\mu_i$.
Its coefficients are used as $w_{i+1}$, the vector of weights that defines the reward function.
A new policy $\pi_{i+1}$ is learned on $R_{i+1}(s)=w_{i+1} \cdot \phi(s)$ and its feature expectation $\mu_{i+1}$ is added to the set of feature expectations used to fit the support vector machine in the following iteration.
The algorithm stops when a stopping criterion is met---for example, after a number of given iterations or when a criterion of similarity between the demonstrated and generated behavior is met.

\begin{algorithm}[t]
\caption{Apprenticeship learning \cite{AbbNg2004icml}}
\label{alg:apprenticeship-learning}
\begin{algorithmic}
\Require $\phi$, $\mu_E$
\State Select a random initial policy $\pi_0$
\State Compute $\mu_0 := \mu(\pi_0)$
\Repeat
  \State Compute $w_{i+1}$ by fitting a SVM on $\mu_E$ and all $\mu_i$
  \State Learn policy $\pi_{i+1}$ on rewards $R_{i+1}(s)=w_{i+1} \cdot \phi(s)$
  \State Compute $\mu_{i+1} := \mu(\pi_{i+1})$
\Until{Stopping criterion met}\\
\Return $w_{i+1}$ as $w^*$
\end{algorithmic}
\end{algorithm}

\section{DESIGNING ROBOT SWARMS BY DEMONSTRATION}
\label{sec:features}

\noindent As shown in Section \ref{sec:related-work}, all demonstration-based methods proposed in swarm robotics so far require that at least some robots exist that can demonstrate the desired behavior.
This clearly prevents the existing approaches from being used to generate new behaviors.
It is our contention that this results from the fact that, in the existing literature, demonstrations have always been conceived as descriptions of \emph{how} the robots should behave.
In this work, we consider demonstrations as descriptions of \emph{what} the swarm should accomplish.
Specifically, we focus here on the class of missions in which what the robots should accomplish is to position themselves in the environment according to a desired distribution.
In this case, a demonstration is a desired final position.
Although this class of missions does not cover all possible missions of interest in swarm robotics, it includes a large share of the missions that have been studied in the literature \cite{GarBir2016weeee,SchUmlSenElm2020FRAI}.

\begin{figure}
    \begin{subfigure}{0.45\linewidth}
    \centering
    \begin{tabular}{c}
    \hline
    \textbf{Sensors}\\ \hline \\[-3mm]
    Promixity\\
    Light\\
    Ground\\
    Range-and-bearing\\
    \\
    \hline
    \textbf{Actuators}\\ \hline \\[-3mm]
    Wheels\\
    \end{tabular}
    \label{tab:table1}
    \end{subfigure}
    \begin{subfigure}{0.5\linewidth}
    \centering
    \includegraphics[width=\textwidth]{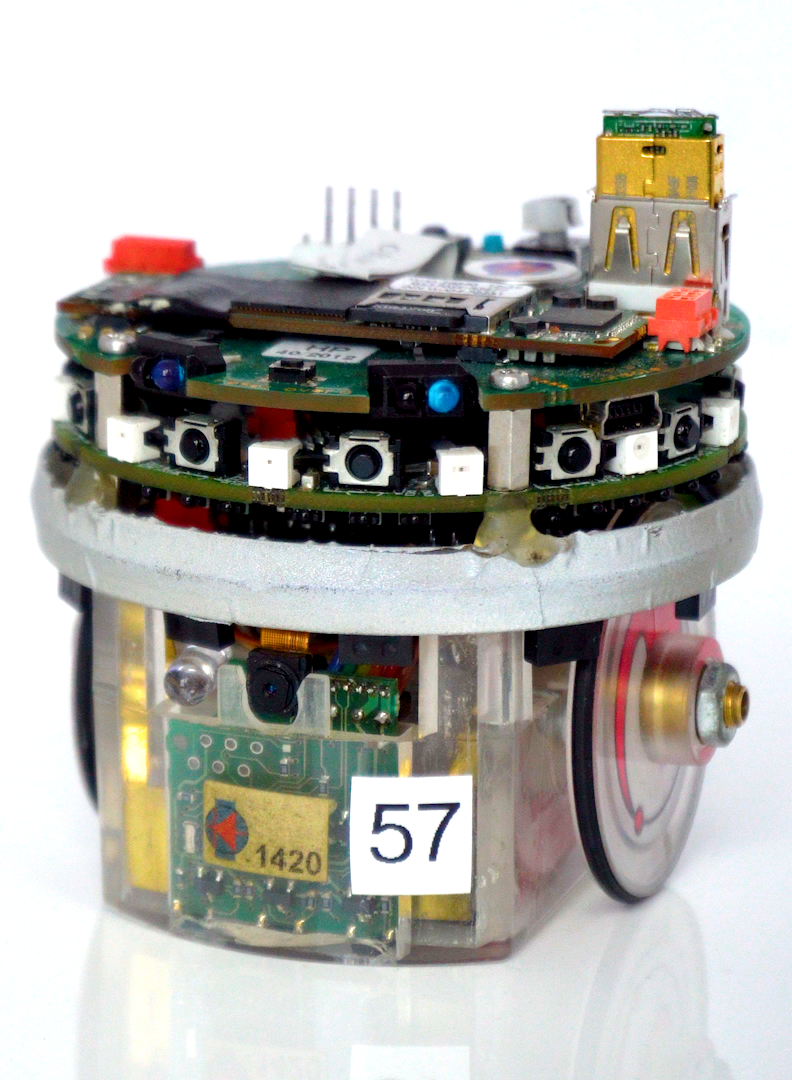}
    \end{subfigure}
    \caption{The \epuck robot and its reference model RM1.1.}
    \label{fig:epuck}
\end{figure}

We propose \DemoCho, an automatic design method that combines apprenticeship learning (see Section \ref{sec:apprenticeship-learning}) with \cho, a state-of-the-art automatic off-line design method to generate control software for robot swarms~\cite{FraBraBru-etal2015SI,HasLigRudBir2021NATUCOM}.
\DemoCho generates control software for the \epuck robot, a two-wheeled robot \cite{MonBonRae-etal2009arsc,GarFraBru-etal2015techrep}, extended by a Linux extension board \cite{LiuWin2011MICPRO} and a range-and-bearing board \cite{GutAlvCam-etal2009icra} (see Figure \ref{fig:epuck}).
Its sensors and actuators were formalized through a reference model, namely RM1.1 \cite{HasLigFra-etal2018techrep}.
According to RM1.1, the robot is endowed with 8 proximity sensors that can perceive obstacles and other robots, 8 light sensors that can perceive a light source, 3 ground sensors that can detect if the floor is white, black or gray, and a range-and-bearing board that provides the number of neighbors perceived and a vector pointing to their center of mass.
The robot is also endowed with two wheels whose velocity can be independently controlled.
We assume that the robots operate in a bounded arena in which the floor is gray and some regions might be white or black.
Outside the arena, there is a light source that is switched on in some missions and off in others.

In \DemoCho, the end user can provide demonstrations of the desired final positions of the robots\footnotemark[2]\footnotetext[2]{See the supplementary material at \url{https://iridia.ulb.ac.be/supp/IridiaSupp2022-003/}}.
\DemoCho then uses the apprenticeship learning algorithm to compute a candidate objective function and \cho to generate control software based on a candidate objective function.
\DemoCho stops after a fixed number of iterations.

Concerning the feature mapping $\phi$, the features we adopted to describe the final position of the robots are based on the distance of each robot from relevant landmarks.
Notably, we consider two classes of landmarks: black or white regions and the nearest peer of each robot.
We scale distances to the interval $[0,1]$ according to $10^{-2 x/d}$ where $d$ is the arena's diameter and $x$ is the distance to the landmark.
Concerning the distance from the regions, if the shortest straight path between the robot and the region is obstructed by a wall, the feature value is set to $0$.
It is worth noting that the set of features is mission-dependent, as the number of black and white regions possibly varies between missions.
Yet, the construction of this mapping is fully automatic and does not require the intervention/analysis of a human expert.
Because all robots of the swarm are interchangeable, the features form an unordered set.
To cast them into a vector in a meaningful way so that the apprenticeship learning algorithm can operate on them, we sort them first by the landmark and then in descending order.
To give an example, in the feature vector $(\phi_{l1,1},\phi_{l1,2},...,\phi_{l1,n},\phi_{l2,1},...)$, $\phi_{l1,1}$ is the feature corresponding to the distance of the nearest robot to landmark $l1$, $\phi_{l1,2}$ is the one corresponding to the distance of the second nearest robot to $l1$, etc.

\section{EXPERIMENTAL SETUP}

\subsection{Design methods}

\noindent To appraise the performance of the control software generated by \DemoCho, we present also the results obtained by \cho and \stick.
\cho designs control software in the form of a probabilistic finite-state machine, assembled from behavioral and conditional modules that are hand-crafted once and for all in a mission-agnostic way \cite{FraBraBru-etal2015SI}.
\stick is an implementation of the classical neuro-evolutionary approach and designs control software in the form of a feed-forward artificial neural network \cite{FraBraBru-etal2014SI}.
Notably, both \cho and \stick require the actual objective function, whereas \DemoCho does not.

\subsection{Missions}

\begin{figure}
    \centering
    \begin{subfigure}[b]{0.45\linewidth}
        \includegraphics[width=\linewidth]{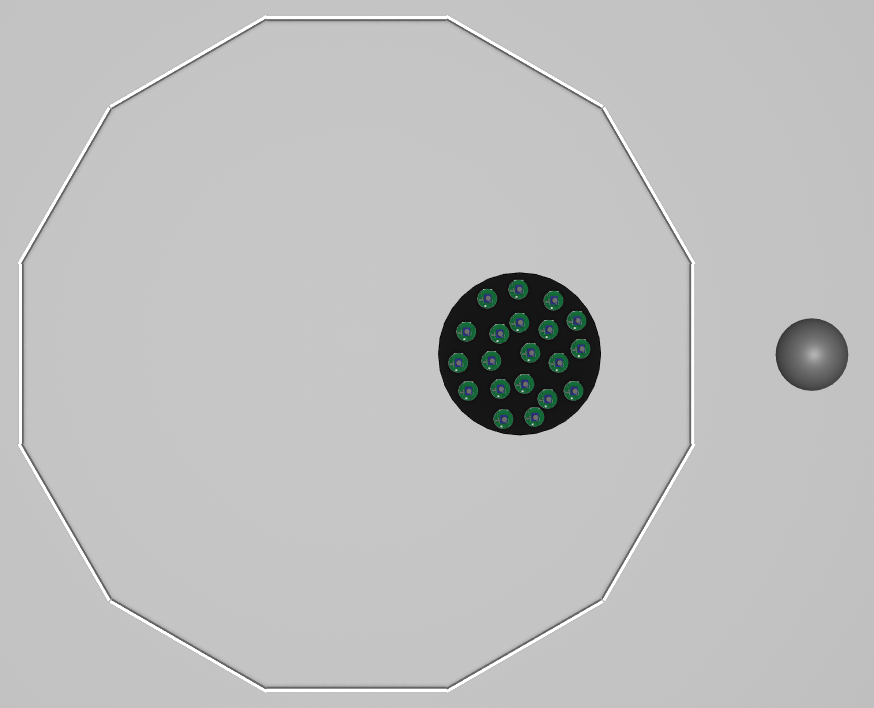}
        \caption{\textsc{Homing}}
        \label{fig:mission-homing}
    \end{subfigure}
        \begin{subfigure}[b]{0.45\linewidth}
        \includegraphics[width=\linewidth]{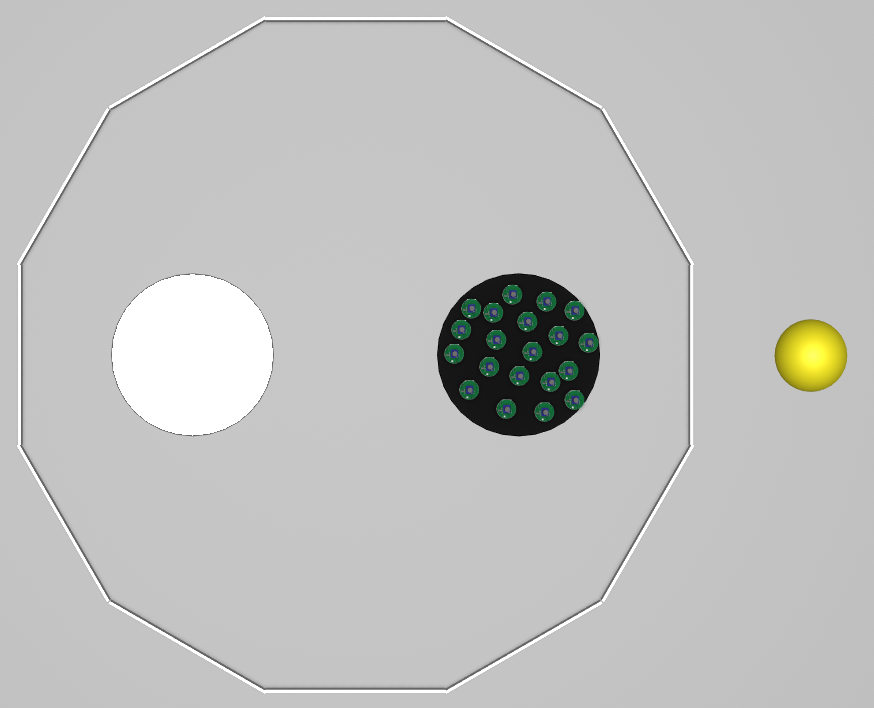}
        \caption{\textsc{AAC}}
        \label{fig:mission-AAC}
    \end{subfigure}\\[2mm]
    \begin{subfigure}[b]{0.45\linewidth}
        \includegraphics[width=\linewidth]{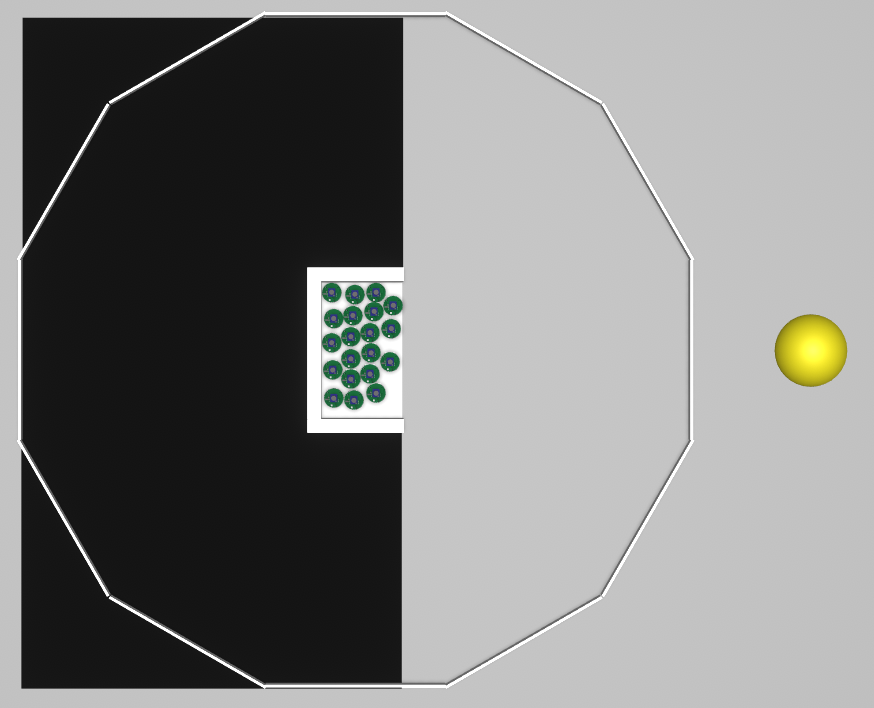}
        \caption{\textsc{SAC}}
        \label{fig:mission-SAC}
    \end{subfigure}
        \begin{subfigure}[b]{0.45\linewidth}
        \includegraphics[width=\linewidth]{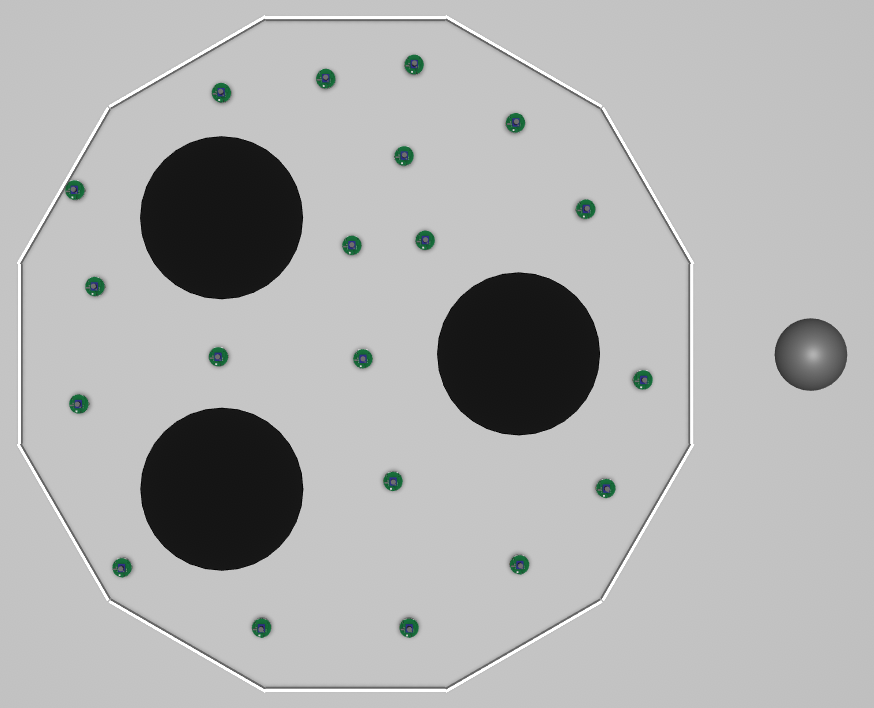}
        \caption{\textsc{CFA}}
        \label{fig:mission-CFA}
    \end{subfigure}
    \caption{Missions and an example of a demonstration.}
    \label{fig:missions}
\end{figure}

\noindent We assess \DemoCho on four missions that were already studied in the literature.
For each of them, an objective function is available because it was defined as part of their specifications in the original works that introduced them.
We report the original objective functions here and we assume that they are accurate representations of the desired collective behaviors.

All mission take place in the same dodecagonal arena of approximately \SI{5}{m^2}.
For all missions, the swarm size is fixed to 20 robots.

In \textsc{Homing} \cite{HasLigRudBir2021NATUCOM}, the swarm must explore the arena and aggregate in the home area represented by a circular black region with radius of \SI{30}{cm} (see Figure \ref{fig:mission-homing}).
The original objective function is $F_\mathit{Homing}=N(T)$, where $N(t)$ is the number of robots in the home area at time $t$ and $T = \SI{180}{\second}$ is the mission duration.

In \textsc{AAC} \cite{FraBraBru-etal2015SI} (aggregation with ambient cues), the swarm must aggregate as quickly as possible in a target area represented by a circular black region with radius of \SI{30}{cm}.
Additionally, the arena contains one white circular region with radius of \SI{30}{cm} and a light source is placed outside of the arena (see Figure \ref{fig:mission-AAC}).
The original objective function is $F_\mathit{AAC}=\sum_{t=1}^{T} N(t)$, where $N(t)$ is the number of robots in the target are at time $t$ and $T=\SI{180}{\second}$ is the mission duration.

In \textsc{SAC} \cite{HasLigBir2022IEEETEC} (shelter with ambient cues), the swarm must aggregate as quickly as possible in a shelter that can only be accessed from one side.
The shelter is indicated by a white rectangular area of \SI{25}{cm} by  \SI{15}{cm} and delimited by three walls, leaving an opening only on one side.
The floor in the arena behind the opening of the shelter is black and a light source is placed outside the arena, facing the open side of the shelter (see Figure \ref{fig:mission-SAC}).
For technical reasons regarding the encoding of the environment in the simulator, the black region is composed by three contiguous rectangular sub-regions, one behind the shelter and one on each of its sides.
The original objective function is $F_\mathit{SAC}=\sum_{t=1}^{T} N(t)$, where $N(t)$ is the number of robots in the shelter at time $t$ and $T=\SI{180}{\second}$ is the mission duration.

In \textsc{CFA} \cite{FraBraBru-etal2015SI} (coverage with forbidden areas), the swarm must spread through the arena while avoiding the forbidden areas represented by three black circular regions with radii of \SI{30}{cm} (see Figure \ref{fig:mission-CFA}).
The original objective function is $E[d(T)]$, the expected distance between a generic point in the arena and the closest robot not on a forbidden area, at the end of T, and $T=\SI{180}{\second}$ is the experiment duration.
To be consistent with the other missions in which the objective function is to be maximized, we reformulate the objective function as $F_\mathit{CFA} = 250 - E[d(T)]$ where $250$ is the theoretical maximum value of $E[d(T)]$.

\subsection{Protocol}

\noindent For each mission, we provided five demonstrations of the final position of the robot swarm to be used by \DemoCho---see the supplementary material\footnotemark[2].
We ran 10 independent design processes for each of the three design methods under analysis.
All design methods adopt the same simulator: \argos \cite{PinTriOgr-etal2012SI}.
\DemoCho was run for 50 iterations, each iteration with a budget of \num{10000} simulation runs per iteration.
\cho and \stick were run with a design budget of \num{10000} simulation runs and optimize the original objective function.
All in all, this grants \DemoCho a budget that is fifty times larger than the one of \cho and \stick.
The goal of this protocol is not to achieve a fair comparison between the three design methods, which could be a rather complex endeavour, see the discussion in Section \ref{sec:conclusions}.
Indeed, \cho and \stick have the clear advantage of being fed with an objective function; the larger budget allocated to \DemoCho is intended to compensate somehow for the fact that \DemoCho has to infer the objective function from the given demonstrations.
In this context, we felt that the primary concern was to provide an appropriate budget to each automatic design process: the one performed by \cho and \stick, and each of the 50 ones performed within each execution of \DemoCho.
Following our previous experience, we allocated to each of these design processes a budget of \num{10000} simulations.
Concerning the choice of the number of iterations to be taken as a stopping criterion for \DemoCho, as no previous literature exist on this issue, we fixed this to a sufficiently large number to make sure that the algorithm had time to converge to a meaningful solution---see the discussion in Section \ref{sec:experimental-results} where we comment \emph{a posteriori} on this choice, in the light of the results obtained through the present study.

We assessed the resulting instances of control software once in simulation and once in reality.
In the experiments with the robots, performance was measured automatically using a tracking system \cite{LegGarKuc-etal2022techrep}.
We provide both a qualitative and a quantitative assessment of the performance of the swarms generated by the three methods under analysis.
The qualitative assessment is based on visual inspection of the generated behaviors.
The quantitative assessment is based on the mission-specific objective function, the same one that \cho and \stick optimize within the design process. 
For a detailed discussion of this choice, we refer the reader to Section \ref{sec:conclusions}.

We report the results in the form of notched boxplots.
In the boxplots, the upper and lower hinges correspond to the first and third quartiles.
The whiskers extend to the largest value of the sample but no further than 1.5 times the interquartile range from the hinge.
Data beyond the whiskers are outliers and are represented by points.
We also report the median of the sample, represented by a line in the box, and a 95\% confidence interval, represented by notches extending from the median line.
If the notches of two boxplots do not overlap, we can conclude that the difference between the medians of the two samples is statistically significant.

The source code, experiment files, and results of all experiments are available as supplementary material\footnotemark[2].

\section{EXPERIMENTAL RESULTS}
\label{sec:experimental-results}

\begin{figure*}[t]
    \centering
    \includegraphics[width=\textwidth]{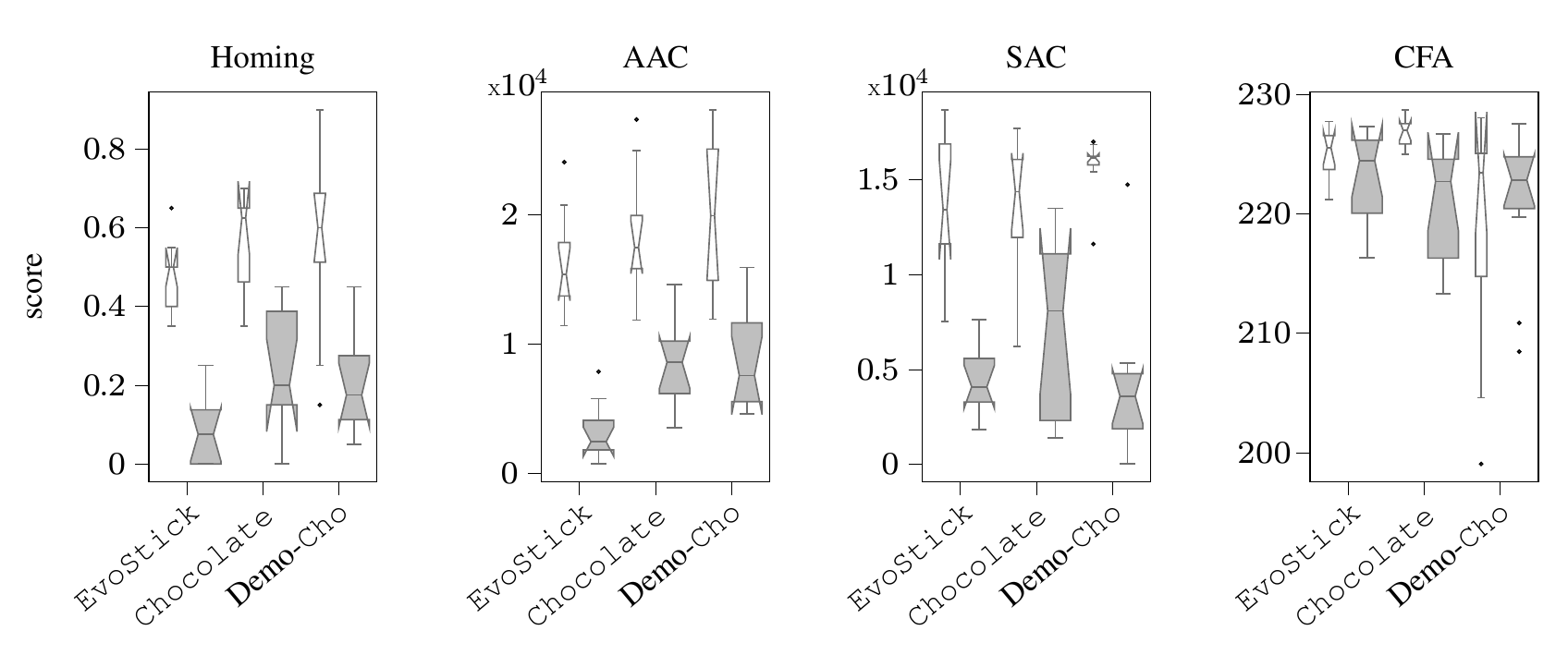}
    \caption{Experimental results obtained in simulation (narrow white boxes) and reality (wide gray boxes).}
    \label{fig:results-boxplot}
\end{figure*}

\noindent Figure \ref{fig:results-boxplot} shows the boxplots of the results obtained in simulation and reality.
The three design methods achieved similar performance in simulation across the four missions, despite the fact that \DemoCho, contrary to \cho and \stick, did not have access to the objective function at design time.
Visual inspection of the generated behaviors in simulation shows that the behaviors generated by \DemoCho match the expectations that one might have on the mission at hand: the robots behave in a meaningful way in all four missions--see supplementary videos\footnotemark[2].
It has to be noted that in the two missions \textsc{AAC} and \textsc{SCA}, the original objective function does not evaluate only the final position---i.e., the one illustrated by the demonstrations provided to \DemoCho---but is computed cumulatively over the whole duration of an experimental run.
Yet, the performance of \DemoCho was not worse than the one of \cho or \stick.

The experiments allowed us to gain some insight on the number of iterations need by \DemoCho to converge to a meaningful solution.
All in all, the selected number of 50 iterations appears to be a reasonably appropriate choice---see the supplementary material\footnotemark[2].
Typically, after the first 10 iterations, the behavior found already reproduces well the given demonstrations.
Further improvement can be observed in the following iterations to then become rare after 40 iterations.
Future work should be devoted to gain a deeper insight in the issue by observing the development of the improvement over an even larger number of iterations.

\begin{figure}
    \centering
    \begin{subfigure}{0.45\linewidth}
        \includegraphics[width=\linewidth]{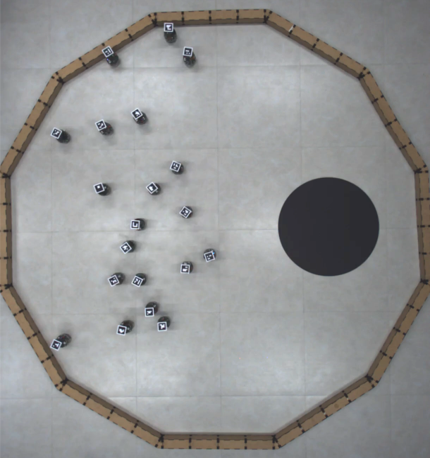}
        \caption{\textsc{Homing}}
        \label{fig:homing-reality}
    \end{subfigure}
    \begin{subfigure}{0.45\linewidth}
        \includegraphics[width=\linewidth]{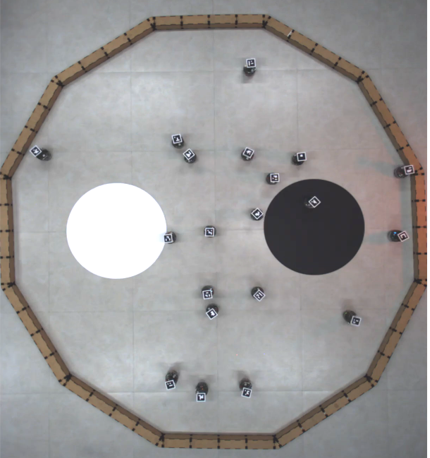}
        \caption{\textsc{AAC}}
        \label{fig:AAC-reality}
    \end{subfigure}\\[2mm]
    \begin{subfigure}{0.45\linewidth}
        \includegraphics[width=\linewidth]{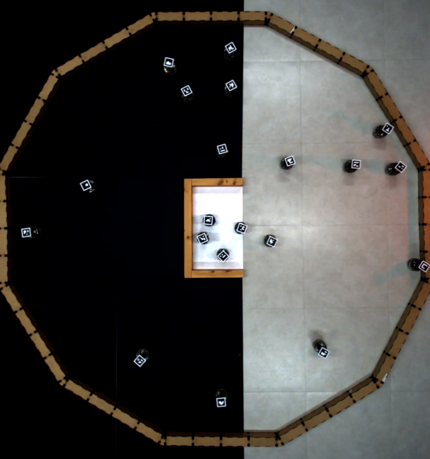}
        \caption{\textsc{SAC}}
        \label{fig:SAC-reality}
    \end{subfigure}
    \begin{subfigure}{0.45\linewidth}
        \includegraphics[width=\linewidth]{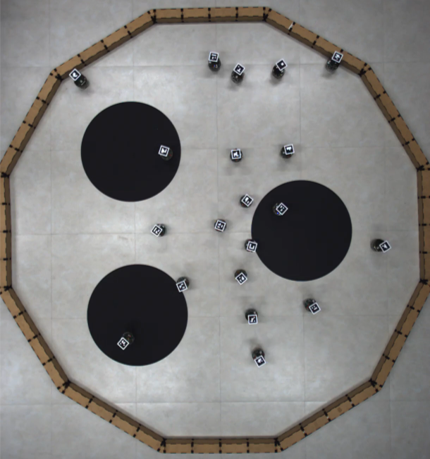}
        \caption{\textsc{CFA}}
        \label{fig:CFA-reality}
    \end{subfigure}
    \caption{The four missions in reality.}
    \label{fig:missions-reality}
\end{figure}

When assessed in reality, all three methods showed a drop in performance---as it is often the case in the automatic design of robot swarms~\cite{LigBir2020SI}.
In the missions \textsc{Homing}, \textsc{SAC}, and \textsc{CFA} the three design methods achieved similar performance in reality.
In \textsc{AAC}, \DemoCho and \cho achieved similar performance in reality and outperformed \stick.
On the basis of these results, we can argue that learning from demonstrations---as opposed to optimizing a given objective function---does not appear to have any major impact on the ability of a modular design method to cross the reality gap. 

\begin{figure}
    \centering
    \includegraphics[width=\linewidth]{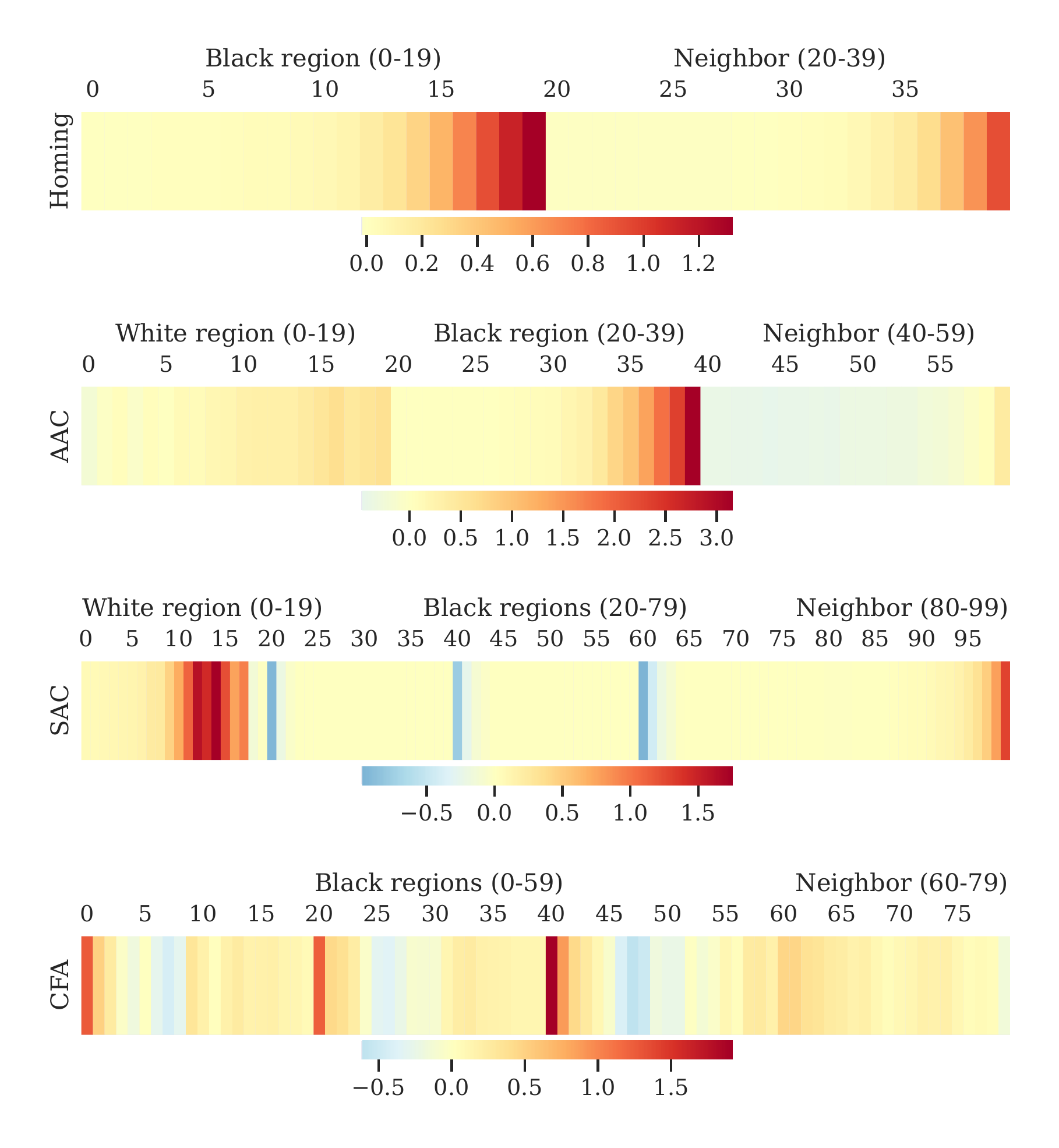}
    \caption{Heat maps of the average weight vectors learned by \DemoCho.}
    \label{fig:w-heatmaps}
\end{figure}

Figure \ref{fig:w-heatmaps} shows the weights $w$ learned by \DemoCho, averaged per mission.
Some general observations can be made for the four missions.
For each group of features---those relating to the same landmark---\DemoCho tends to put larger weights on the feature of lower value, that is those corresponding to the robots that are the farthest from the landmark.
Indeed, minimizing the distance of the farthest robots also guarantees that the distance of all robots is minimized.
When looking at the weights for the specific missions, we can observe the following:
In \textsc{Homing}, the distance to the black region was selected by \DemoCho as the most important feature.
Albeit to a lesser extent, the distance to the nearest neighbor was considered important as well.
Thus, the design process rewarded behaviors that aggregate tightly in the home area.
Also in \textsc{AAC}, \DemoCho selects the distance to the black region as the most important feature.
Unlike in \textsc{Homing}, however, the distance to the nearest neighbor was not considered important; neither was the one to the white region.
For this mission, the design process rewarded behaviors that aggregate in the target area.
The tightness of the aggregation possibly resulted implicitly, as all robots must fit in the target area.
In \textsc{SAC}, the design process selected two important features: the distance to the white region and the one to the nearest peer.
The selection of these two features can be interpreted to describe an aggregation behavior in the shelter.
Curiously, unlike for the other features, \DemoCho assigned the highest weight to the feature associated with the sixth farthest robot from the white region, rather than the feature associated with the farthest one.
This might be explained by the fact that it is unlikely that all the robots eventually reach the shelter and five robots outside the shelter at the end of the experimental run is a common outcome.
Additionally, we observe three features that \DemoCho penalizes through the assignment of a negative weight: the distance of the nearest robot to each of the black regions.
Maximizing the distance between the nearest robot and a landmark guarantees that the distance of all robots is maximized.
In \textsc{CFA}, \DemoCho selected three groups of features as important: the distance to each of the black regions.
In this case, the weights were selected to favor the presence of the robots nearby each of the black regions: the highest weight is associated with the feature corresponding to the distance of robot closest to the landmark.
Additionally, \DemoCho slightly penalizes the features corresponding to the distances from the landmark of the fifth to eighth nearest robots.
As a result, the design process aimed to keep the robots close to the forbidden areas without favoring an aggregation.
Additionally, some importance is placed on the features describing the inter-robot distance: a slightly positive weight is associated to the distance of nearest peers.

The interpretation for the weights is particularly straightforward for \textsc{Homing}, \textsc{AAC}, and \textsc{SAC}, while it is less intuitive for \textsc{CFA}.
Indeed, in \textsc{CFA}, one could have expected more emphasis on the inter-robot distance and the penalization of the distance to the forbidden areas.
Nonetheless, excluding two outliers, the performance achieved by \DemoCho in this mission is satisfactory and the behavior of the robots appears to be meaningful at visual inspection---see supplementary videos\footnotemark[2].


\section{CONCLUSIONS}
\label{sec:conclusions}

\noindent In this work, we have presented \DemoCho, an automatic method for designing control software of robot swarms that combines inverse reinforcement learning with automatic modular design.
Instead of optimizing an explicitly defined objective function, \DemoCho generates control software based on provided demonstrations.
In our experiments, \DemoCho was able to create satisfactory behaviors to perform four missions that were previously studied in the literature.
Expressing a desired outcome in terms of a mathematical function is unintuitive and requires the attention of an expert.
Specifying desired behaviors through demonstrations is natural and intuitive and could allow even end users without any technical expertise to specify their desired behaviors.

In the experiments presented in this paper, we accept the original assumption made by the proponents of the missions that the objective function accurately specifies the desired behavior.
We therefore use this objective function for the final assessment of the behaviors produced by \DemoCho on the basis of the given demonstrations.
However, this way of assessing performance is viable only for missions that already have been specified via the definition of an objective function.
A general protocol to assess behaviors generated from demonstrations could be defined on the basis of an appropriate metric that measures the degree of similarity between the given demonstrations and the generated behavior.
Yet, the goal would not be to reproduce the demonstrations but to generalize with respect to them.
An appropriate protocol could take take inspiration from the classical cross-validation and leave-one-out procedures typically adopted in machine learning.

A protocol should also be defined to compare in fair way methods based on demonstrations with traditional ones that optimize a given objective function.
The latter clearly have an advantage on the former, which have to infer an objective function from the given examples.
An appropriate protocol should test also traditional methods on an objective function other than the one they used at design time.
For example, two experts might define one objective function each.
One of these objective functions could be used by the traditional methods in the design phase; and the other could be used to test both traditional methods and demonstration-based ones.
This would put the two methods on the same foot for what concerns the evaluation.

In the future, we will extend \DemoCho to missions that can be represented through the final position of elements other than the robots---e.g., objects to be clustered, gathered, spread in the environment.
Additionally, we will investigate the minimum number of demonstrations necessary to design a desired behavior and more generally, the impact the number of demonstrations and their diversity have on the quality of the behaviors that can be obtained.




%

\bibliographystyle{IEEEtran}
\bibliography{demiurge-bib/definitions,demiurge-bib/author,demiurge-bib/address,demiurge-bib/proceedings-short,demiurge-bib/journal-short,demiurge-bib/publisher,demiurge-bib/series-short,demiurge-bib/institution,demiurge-bib/bibliography,additions}

\end{document}